\documentclass{article}

\usepackage[nonatbib,final]{nips_2017}
\usepackage[utf8]{inputenc} 
\usepackage[T1]{fontenc}    
\usepackage[colorlinks]{hyperref}       
\usepackage{url}            
\usepackage{booktabs}       
\usepackage{amsfonts}       
\usepackage{nicefrac}       
\usepackage{microtype}      
\usepackage{graphicx}
\usepackage{multirow}
\usepackage{changepage}
\usepackage{array}
\usepackage{comment}
\usepackage{enumitem}

\title{CURE-TSR: Challenging Unreal and Real Environments for Traffic Sign Recognition}

\author{
  Dogancan~Temel, Gukyeong~Kwon
  \thanks{Equal contribution.}, Mohit~Prabhushankar\footnotemark[1], and Ghassan~AlRegib\\
  Georgia Institute of Technology\\
  Center for Signal and Information Processing \\
  \texttt{\{cantemel, gukyeong.kwon, mohit.p, alregib\}@gatech.edu} \\
}

\begin{document}

\begin{description}[labelindent=-1cm,leftmargin=1cm,style=multiline,labelsep=2em]

\item[\textbf{Citation}]{D. Temel, G. Kwon*, M. Prabhushankar*, and G. AlRegib, “CURE-TSR: Challenging unreal and real environments for traffic sign recognition,” in Neural Information Processing Systems (NIPS) Workshop on Machine Learning for Intelligent Transportation Systems (MLITS),  Long Beach, U.S., December 2017.} \\ \\


\item[\textbf{Review}]{Date of acceptance to NIPS MLITS : November 14 2017} \\ \\

\item[\textbf{Data/Codes}]{\url{https://ghassanalregib.com/cure-tsr/}}
\\

\item[\textbf{Bib}]  {@INPROCEEDINGS\{Temel2017\_NIPS,\\
author=\{D. Temel and G. Kwon and M. Prabhushankar and G. AlRegib\},\\ 
booktitle=\{Neural Information Processing Systems (NIPS) Workshop on Machine Learning for Intelligent Transportation Systems (MLITS)\},\\
title=\{CURE-TSR: Challenging unreal and real environments for traffic sign recognition\}, \\
year=\{2017}\},\\
month=\{December\},
\} \\



\item[\textbf{Contact}]{\href{mailto:alregib@gatech.edu}{alregib@gatech.edu}~~~~~~~\url{https://ghassanalregib.com/}\\ \href{mailto:dcantemel@gmail.com}{dcantemel@gmail.com}~~~~~~~\url{http://cantemel.com/}}
\end{description}

\newpage
\clearpage

\maketitle

\begin{abstract}
In this paper, we investigate the robustness of traffic sign recognition algorithms under challenging conditions. Existing datasets are limited in terms of their size and challenging condition coverage, which motivated us to generate the Challenging Unreal and Real Environments for Traffic Sign Recognition (\href{https://ghassanalregib.com/cure-tsr/}{CURE-TSR}) dataset. It includes more than two million traffic sign images that are based on real-world and simulator data. We benchmark the performance of existing solutions in real-world scenarios and analyze the performance variation with respect to challenging conditions. We show that challenging conditions can decrease the performance of baseline methods significantly, especially if these challenging conditions result in loss or misplacement of spatial information. We also investigate the effect of data augmentation and show that utilization of simulator data along with real-world data enhance the average recognition performance in real-world scenarios. The dataset is publicly available at  \href{https://ghassanalregib.com/cure-tsr/}{https://ghassanalregib.com/cure-tsr/}.
\end{abstract}

\section{Introduction}
Autonomous vehicles are transforming existing transportation systems. As we step up the ladder of autonomy, more critical functions are performed by algorithms, which demands more robustness. In case of following traffic rules, robust sign recognition systems are essential unless we have prior information about traffic sign types and locations. It is a common practice to test the robustness of these systems with traffic datasets
\cite{Grigorescu2003,Timofte2009,Timofte2014,Belaroussi2010,Larsson2011,Stallkamp2011,Stallkamp2012,Houben2013,Mogelmose2012,Zhu2016}. However, majority of these datasets are limited in terms of challenging environmental conditions. There is usually no metadata corresponding to challenging conditions or levels in these datasets, which are also limited in terms of dataset size. Moreover, the relationship between challenging conditions and algorithmic performance is not analyzed in these studies. Lu \textit{et al.} \cite{Lu2017} investigated the traffic sign detection performance with respect to challenging adversarial examples and showed that adversarial perturbations are effective only in specific situations. Das \textit{et al.} \cite{Das2017} showed the vulnerabilities of existing systems and suggested JPEG compression to eliminate adversarial effects. Even though both of these studies analyze algorithmic performance variation with respect to specific challenging situations, adversarial examples are inherently different from realistic challenging scenarios.

In this paper, we investigate the traffic sign recognition performance of commonly used methods under realistic challenging conditions. To eliminate the shortcomings of existing datasets, we introduce the Challenging Unreal and Real Environments for Traffic Sign Recognition (\texttt{CURE-TSR}) dataset. The contributions of this paper are $4$ folds.

\begin{itemize}
    \item We introduce the most comprehensive publicly-available traffic sign recognition dataset with controlled challenging conditions.
    \item We provide a detailed analysis of the benchmarked algorithms in terms of their recognition performance under challenging conditions. Based on this analysis, we identify the vulnerabilities of algorithms with respect to challenging conditions, which should give insights into the use of such models under certain conditions.
    \item We provide images that originate from captured sequences as well as synthesized sequences, which would lead to a better understanding of similarities/differences between real-world and simulator data in terms of algorithmic performance. This understanding can be utilized to generate more realistic datasets and minimize the need for real-world data collection that requires significant resources.
    \item We use diverse data augmentation methods and show that utilization of limited simulator data along with real-world data can enhance the recognition performance. This observation shows that simulated environments can enhance the performance of data-driven methods in real-world scenarios even when there is a difference between target and source domains.
\end{itemize}







\section{Dataset}
Timofte \textit{et al.} \cite{Timofte2014} introduced the Belgium traffic sign classification (BelgiumTSC) dataset whose images were acquired with a van that had 8 roof-mounted cameras. Acquisition vehicle cruised in streets of Belgium and images were captured every meter. A subset of these images were selected and traffic signs were cropped to obtain the BelgiumTSC dataset. Stallkamp \textit{et al.} \cite{Stallkamp2011,Stallkamp2012} introduced the German traffic sign recognition benchmark (GTSRB) dataset, which was acquired during daytime in Germany. Each traffic sign instance in the dataset is adjusted to have $30$ images. BelgiumTSC and GTSRB datasets are limited in terms of challenging environmental conditions and they do not include metadata related to the type of challenging conditions or their levels. Because of limited control in data acquisition setup, it is not possible to perform controlled experiments with these datasets. The total number of annotated signs including BelgiumTSC and GTSRB datasets is around $60,000$, which may not be sufficient to test the robustness of recognition algorithms comprehensively. To compensate the shortcomings in the literature, we introduce the \texttt{CURE-TSR} dataset.
Main characteristics of BelgiumTSC, GTSRB, and \texttt{CURE-TSR} datasets are summarized in Table \ref{tab_datasets}.

\begin{table*}[htbp!]
\scriptsize
\centering
\caption{Main characteristics of BelgiumTSC, GTSRB, and CURE-TSR datasets.}
\label{tab_datasets}
\begin{tabular}{|c|c|c|c|c|c|c|}
\hline
 \textbf{\begin{tabular}[p]{@{}c@{}}Dataset  \end{tabular}} & \textbf{\begin{tabular}[c]{@{}c@{}}Number of \\images\end{tabular}} & \textbf{\begin{tabular}[c]{@{}c@{}}Number of \\annotated\\ images\end{tabular}}   & \textbf{\begin{tabular}[c]{@{}c@{}}Number of \\ sign types\end{tabular}} &\textbf{\begin{tabular}[c]{@{}c@{}}Sign \\ size\end{tabular}}&\textbf{\begin{tabular}[c]{@{}c@{}}Origin of \\ the videos\end{tabular}}&\textbf{\begin{tabular}[c]{@{}c@{}}Acquisition \\ device\end{tabular}} \\ \hline

\begin{tabular}[c]{@{}c@{}}\textbf{BelgiumTSC} \\ \cite{BelgiumTS_dataset}\end{tabular}    &\begin{tabular}[c]{@{}c@{}}7,095  - \\ 7,125\end{tabular} & \begin{tabular}[c]{@{}c@{}}All \\images\end{tabular}   &62  &\begin{tabular}[c]{@{}c@{}}11x10  to\\ 562x438  \end{tabular}& \begin{tabular}[c]{@{}c@{}}Captured \\in\\ Belgium\end{tabular}&\begin{tabular}[c]{@{}c@{}}Color\\ cameras\end{tabular}\\ \hline

\begin{tabular}[c]{@{}c@{}}\textbf{GTSRB} \\ \cite{GTS_dataset}\end{tabular}    &\begin{tabular}[c]{@{}c@{}}133,000  - \\ 144,769\end{tabular} & 51,840   &43  &\begin{tabular}[c]{@{}c@{}}15x15 to\\ 250x250\end{tabular} & \begin{tabular}[c]{@{}c@{}}Captured \\in\\ Germany\end{tabular}& \begin{tabular}[c]{@{}c@{}}Prosilica GC\\ 1380CH\\ color camera\end{tabular} \\ \hline

\begin{tabular}[c]{@{}c@{}}\textbf{CURE-TSR} \\ \cite{curetsr_dataset}\end{tabular} &2,206,106 &\begin{tabular}[c]{@{}c@{}}All \\images\end{tabular}   & 14 &\begin{tabular}[c]{@{}c@{}}3x7 to \\ 206x277\end{tabular}& \begin{tabular}[c]{@{}c@{}}Captured in \\ Belgium and \\ Generated in \\Unreal Engine 4\end{tabular} &\begin{tabular}[c]{@{}c@{}}Color\\ cameras\end{tabular}  \\ \hline

\end{tabular}
\end{table*}

\begin{figure}[htbp!]
\centering
\begin{minipage}[b]{0.49\linewidth}
  \centering
\includegraphics[width=\linewidth, trim= 10mm 5mm 10mm 5mm]{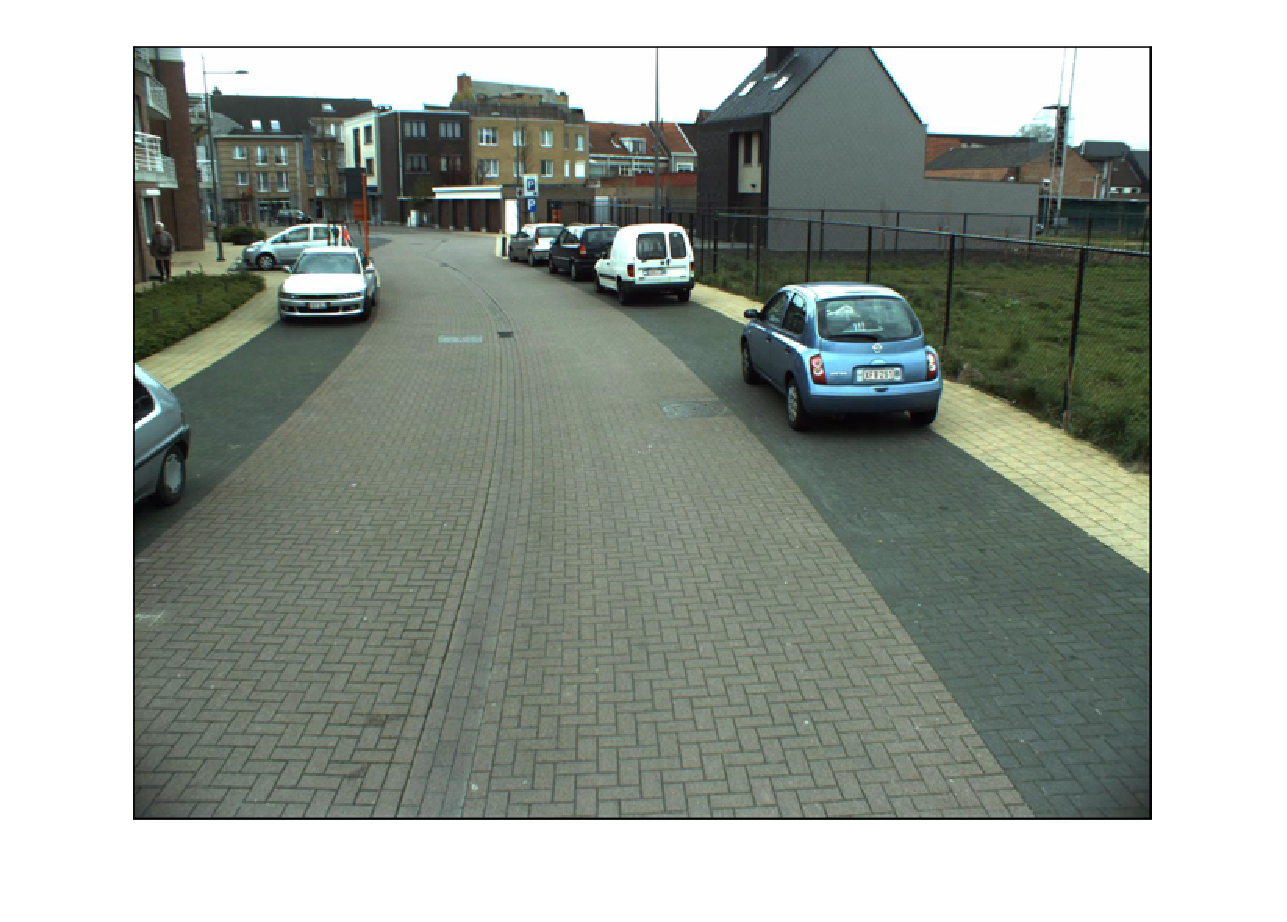}
  \vspace{0.01cm}
  \centerline{\footnotesize{(a) Real-world (real) image }}
\end{minipage}
\begin{minipage}[b]{0.49\linewidth}
  \centering
\includegraphics[width=\linewidth, trim= 10mm 5mm 10mm 5mm]{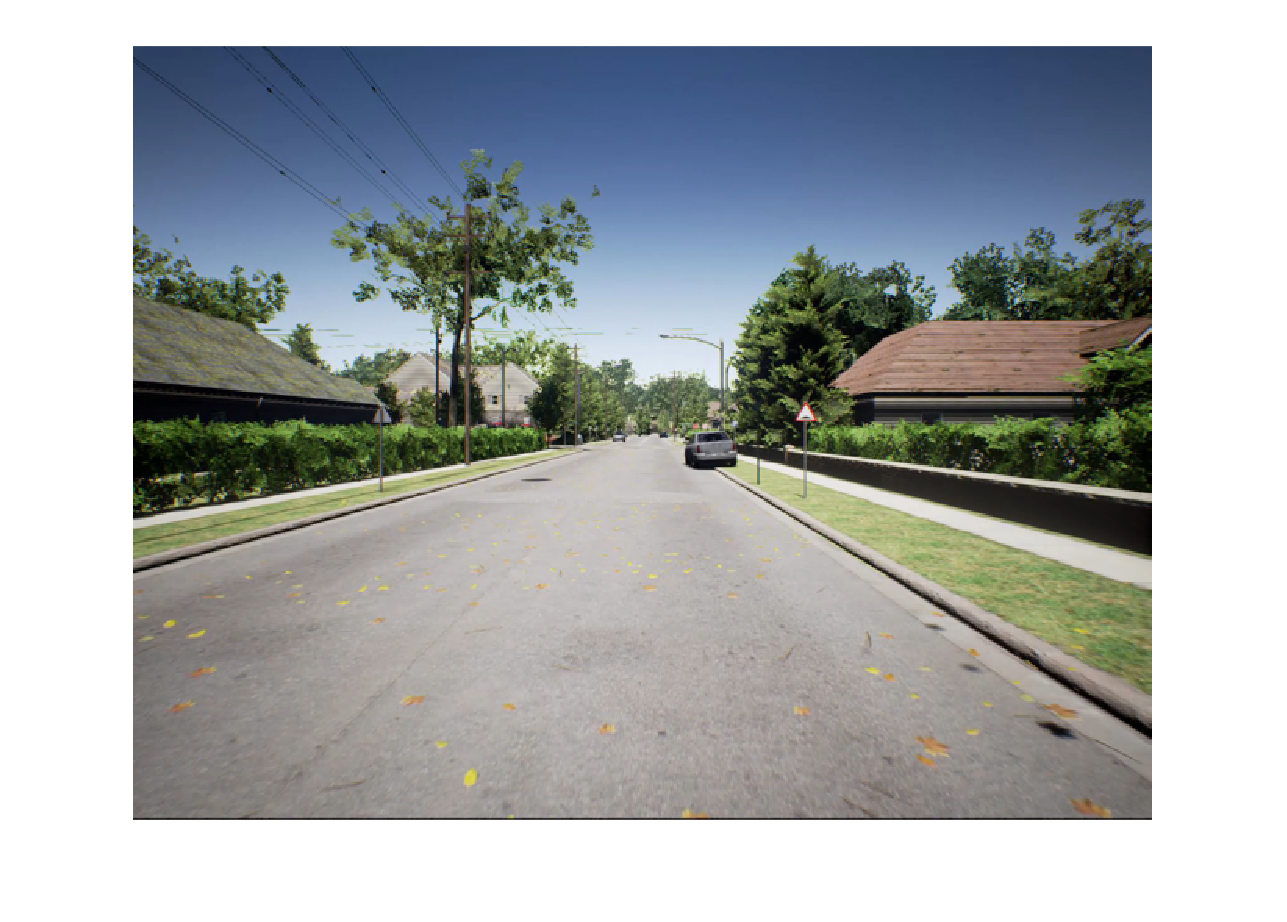}
  \vspace{0.01cm}
  \centerline{\footnotesize{(b) Simulator (unreal) image }}
\end{minipage}
\
\caption{Real and unreal environments.}
\label{fig:scene}
\end{figure}

\begin{center}
\begin{figure*}[!h]
\centering
\setlength{\tabcolsep}{0.2 em}
\scriptsize
\renewcommand{\arraystretch}{0.5}
\begin{tabular}{cccccccccccccc}
\includegraphics[width=0.06\linewidth,height=0.06\linewidth]{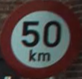} &
\includegraphics[width=0.06\linewidth,height=0.06\linewidth]{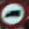} &
\includegraphics[width=0.06\linewidth,height=0.06\linewidth]{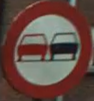} &
\includegraphics[width=0.06\linewidth,height=0.06\linewidth]{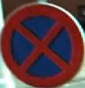} & \includegraphics[width=0.06\linewidth,height=0.06\linewidth]{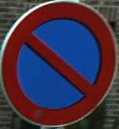} &
\includegraphics[width=0.06\linewidth,height=0.06\linewidth]{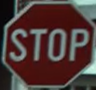} &
\includegraphics[width=0.06\linewidth,height=0.06\linewidth]{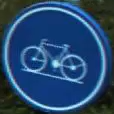} & \includegraphics[width=0.06\linewidth,height=0.06\linewidth]{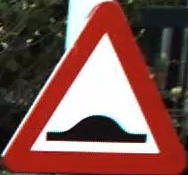} &
\includegraphics[width=0.06\linewidth,height=0.06\linewidth]{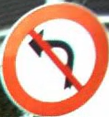} &
\includegraphics[width=0.06\linewidth,height=0.06\linewidth]{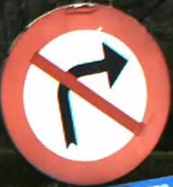} &
\includegraphics[width=0.06\linewidth,height=0.06\linewidth]{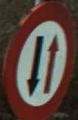} & \includegraphics[width=0.06\linewidth,height=0.06\linewidth]{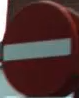} &
\includegraphics[width=0.06\linewidth,height=0.06\linewidth]{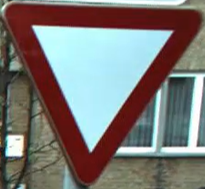} &
\includegraphics[width=0.06\linewidth,height=0.06\linewidth]{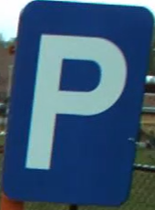} \\
\includegraphics[width=0.06\linewidth,height=0.06\linewidth]{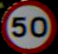} &
\includegraphics[width=0.06\linewidth,height=0.06\linewidth]{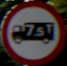} &
\includegraphics[width=0.06\linewidth,height=0.06\linewidth]{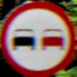} &
\includegraphics[width=0.06\linewidth,height=0.06\linewidth]{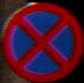} & \includegraphics[width=0.06\linewidth,height=0.06\linewidth]{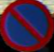} &
\includegraphics[width=0.06\linewidth,height=0.06\linewidth]{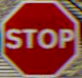} &
\includegraphics[width=0.06\linewidth,height=0.06\linewidth]{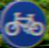} & \includegraphics[width=0.06\linewidth,height=0.06\linewidth]{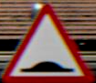} &
\includegraphics[width=0.06\linewidth,height=0.06\linewidth]{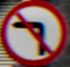} &
\includegraphics[width=0.06\linewidth,height=0.06\linewidth]{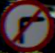} &
\includegraphics[width=0.06\linewidth,height=0.06\linewidth]{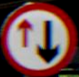} & \includegraphics[width=0.06\linewidth,height=0.06\linewidth]{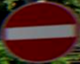} &
\includegraphics[width=0.06\linewidth,height=0.06\linewidth]{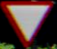} &
\includegraphics[width=0.06\linewidth,height=0.06\linewidth]{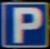} \\
\tiny{speed} & \tiny{goods} & \tiny{no} & \tiny{no} & \tiny{no} & \multirow{2}{*}{\tiny{stop}} & \multirow{2}{*}{\tiny{bicycle}} & \multirow{2}{*}{\tiny{hump}} & \tiny{no} & \tiny{no} & \tiny{priority} & \tiny{no} & \multirow{2}{*}{\tiny{yield}} & \multirow{2}{*}{\tiny{parking}} \\
\tiny{limit} & \tiny{vehicles} & \tiny{overtaking} & \tiny{stopping} & \tiny{parking} & & & & \tiny{left} & \tiny{right} & \tiny{to} & \tiny{entry} & & \\
\end{tabular}
\caption{Traffic signs in real ($1^{st}$ row) and unreal ($2^{nd}$ row) environments.}
\label{fig: sign_type}
\end{figure*}
\end{center}

Traffic sign images in the \texttt{CURE-TSR} dataset were cropped from the \texttt{CURE-TSD} dataset \cite{curetsd_dataset}, which includes around $1.7$ million real-world and simulator images with more than $2$ million traffic sign instances. Real-world images were obtained from the BelgiumTS video sequences and simulated images were generated with the Unreal Engine 4 game development tool. In Fig. \ref{fig:scene}, we show a sample real-world image and a simulator image. In the rest of this paper, we refer to simulator generated images as unreal images and real-world images as real images. As observed in sample images, both real and unreal images are usually from urban environments. While deciding on the type of traffic signs to be included in real and unreal sequences, we focused on two main criteria. First, not every sign type can be reasonably located in unreal sequences. Second, there are limited number of common signs between the package utilized in the generation of unreal sequences and real sequences. Based on the aforementioned selection criteria, we narrowed down number of traffic signs to 14 types as shown in Fig. \ref{fig: sign_type}. Sign types include \textit{speed limit}, \textit{goods vehicles}, \textit{no overtaking}, \textit{no stopping}, \textit{no parking}, \textit{stop}, \textit{bicycle}, \textit{hump}, \textit{no left}, \textit{no right}, \textit{priority to}, \textit{no entry}, \textit{yield}, and \textit{parking}.

\begin{center}
\begin{figure*}[!htbp]
\centering
\setlength{\tabcolsep}{0.25 em}
\scriptsize
\renewcommand{\arraystretch}{0.5}
\begin{tabular}{ccccccccccccc}
\includegraphics[width=0.06\linewidth,height=0.06\linewidth]{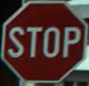}  &
\includegraphics[width=0.06\linewidth,height=0.06\linewidth]{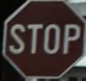} &
\includegraphics[width=0.06\linewidth,height=0.06\linewidth]{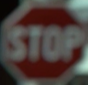} &
\includegraphics[width=0.06\linewidth,height=0.06\linewidth]{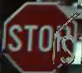} &
\includegraphics[width=0.06\linewidth,height=0.06\linewidth]{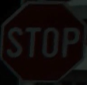} &
\includegraphics[width=0.06\linewidth,height=0.06\linewidth]{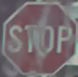} &
\includegraphics[width=0.06\linewidth,height=0.06\linewidth]{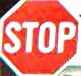} &
\includegraphics[width=0.06\linewidth,height=0.06\linewidth]{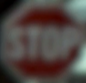} &
\includegraphics[width=0.06\linewidth,height=0.06\linewidth]{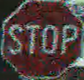} &
\includegraphics[width=0.06\linewidth,height=0.06\linewidth]{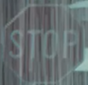} & \includegraphics[width=0.06\linewidth,height=0.06\linewidth]{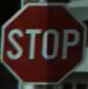} &
\includegraphics[width=0.06\linewidth,height=0.06\linewidth]{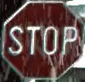} &
\includegraphics[width=0.06\linewidth,height=0.06\linewidth]{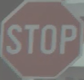} 
\\
\includegraphics[width=0.06\linewidth,height=0.06\linewidth]{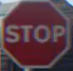} &
\includegraphics[width=0.06\linewidth,height=0.06\linewidth]{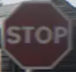} &
\includegraphics[width=0.06\linewidth,height=0.06\linewidth]{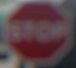} &
\includegraphics[width=0.06\linewidth,height=0.06\linewidth]{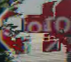} &
\includegraphics[width=0.06\linewidth,height=0.06\linewidth]{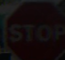} &
\includegraphics[width=0.06\linewidth,height=0.06\linewidth]{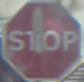} & 
\includegraphics[width=0.06\linewidth,height=0.06\linewidth]{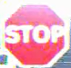} &
\includegraphics[width=0.06\linewidth,height=0.06\linewidth]{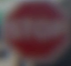} &
\includegraphics[width=0.06\linewidth,height=0.06\linewidth]{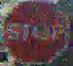} &
\includegraphics[width=0.06\linewidth,height=0.06\linewidth]{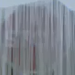} & \includegraphics[width=0.06\linewidth,height=0.06\linewidth]{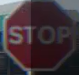} &
\includegraphics[width=0.06\linewidth,height=0.06\linewidth]{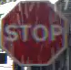} &
\includegraphics[width=0.06\linewidth,height=0.06\linewidth]{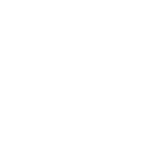} \\
\tiny{No} & \tiny{Decolor-} & \tiny{Lens} & \tiny{Codec} & \multirow{2}{*}{\tiny{Darkening}} & \tiny{Dirty} & \multirow{2}{*}{\tiny{Exposure}} & \tiny{Gaussian} & \multirow{2}{*}{\tiny{Noise}} & \multirow{2}{*}{\tiny{Rain}} & \multirow{2}{*}{\tiny{Shadow}} & \multirow{2}{*}{\tiny{Snow}} & \multirow{2}{*}{\tiny{Haze}} \\
\tiny{Challenge} & \tiny{ization} & \tiny{Blur} & \tiny{Error} & & \tiny{Lens} & & \tiny{Blur} & & & & & \\
\end{tabular}
\caption{Stop signs under challenging conditions in real ($1^{st}$ row) and unreal ($2^{nd}$ row) environments.}
\label{fig: sign_distortion}
\end{figure*}

\end{center}

Unreal and real sequences were processed with state-of-the-art visual effect software Adobe(c) After Effects to simulate challenging conditions, which include \texttt{rain}, \texttt{snow}, \texttt{haze}, \texttt{shadow}, \texttt{darkness}, \texttt{brightness}, \texttt{blurriness}, \texttt{dirtiness}, \texttt{colorlessness}, \texttt{sensor} and \texttt{codec errors}. The key component in this study is not the number of traffic signs but the number of challenging conditions and the context of each traffic sign in a virtual dataset and its corresponding real dataset. If one considers a traffic sign in a challenging condition as a distinct configuration, then we end up with 182 (14x13) distinct configurations in real sequences and 168 (14x12) distinct configurations in virtual sequences. In Fig. \ref{fig: sign_distortion}, we show sample stop sign images under challenging conditions in both real and unreal environments. We included \texttt{codec error} as an edge case to test the limits of benchmarked methods. Recognizing traffic signs with \texttt{codec errors} can be challenging even for subjects because of significant misalignment. If a sign is totally misaligned, it will not be possible to recognize it at cropped location but in case there is residual, it can still be possible to recognize that traffic sign. \texttt{Codec-related errors} can be critical in various applications including but not limited to remote driving. Overall, there are 5 challenge levels for each challenge category, which are shown in Appendix~\ref{sec:AppendixA}.


\vspace{-6.0mm}

\section{Experiments}
\subsection{Baseline Methods, Dataset, and Performance Metric}
\label{exp_setup}
In the German traffic sign recognition benchmark (GTSRB) \cite{Stallkamp2011}, histogram of oriented gradient (HOG) features were utilized to report the baseline results. In the Belgium traffic sign classification (BelgiumTSC) benchmark, cropped traffic sign images were converted into grayscale and rescaled to $28 \times 28$ patches, which were included in the baseline. Moreover, HoG features were also used as a baseline method. They classified traffic sign images with methods including support vector machines (SVMs). Similar to GTSRB and BelgiumTSC datasets, we use rescaled grayscale and color images as well as HoG features as baseline. In the final classification stage, we utilize one-vs-all SVMs with radial basis kernels and softmax classifiers. In addition to aforementioned techniques, we also use a shallow convolutional neural network, which consists of two convolutional layers followed by two fully connected layers, and a softmax classifier. We preprocessed images using $l_2$ normalization, mean subtraction, and division by standard deviation.  

Traffic sign images originate from $49$ video sequences, which are split into approximately $70\%$ training set and $30\%$ test set. Video sequences were split one sign at a time, starting from the least common sign. Once video sequences were assigned to training or testing sets, splitting continued from the remaining sequences until all the sequences were classified. In the first experiment set, we utilize $7,292$ traffic sign images in the training stage obtained from challenge-free real training sequences. In the testing, we utilize $3,334$ images from each challenge category and level, which adds up to $200,040$ images ($3,334$ images $\times 12$ challenge types $\times 5$ levels).
As performance metric, we utilize classification accuracy, which corresponds to the percentage of traffic signs that are correctly classified.

\subsection{Experiment 1: Recognition in Real Environments under Challenging Conditions}
\label{exp1}
\begin{center}
\begin{figure}[htbp!]
\begin{minipage}[b]{0.23\linewidth}
  \centering
\includegraphics[width=\textwidth, trim = 20mm 60mm 20mm 60mm]{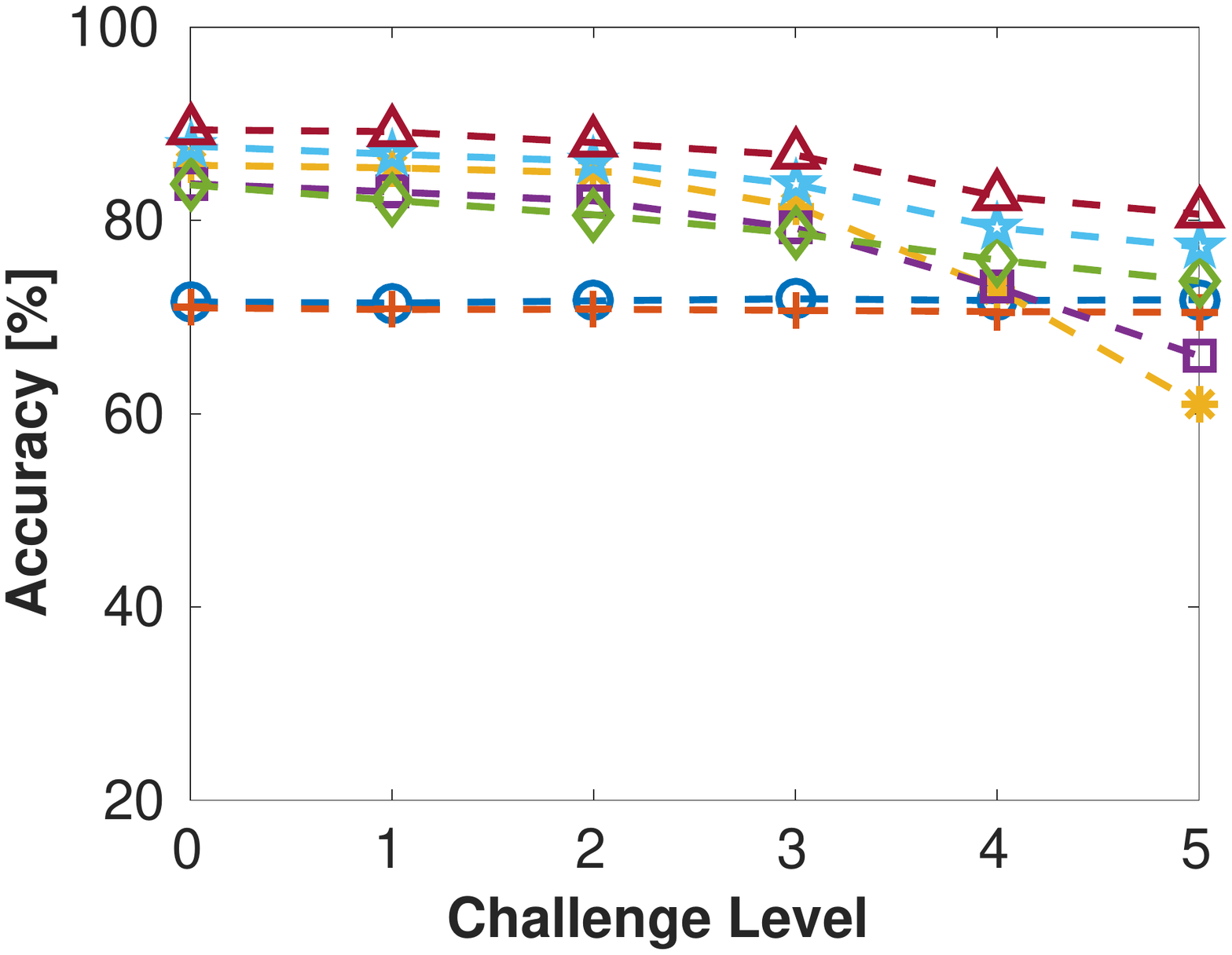}
  \vspace{0.01cm}
  \centerline{\footnotesize{(a)Decolorization}}

\end{minipage}
\hfill
\begin{minipage}[b]{0.23\linewidth}
  \centering
\includegraphics[width=\textwidth, trim = 20mm 60mm 20mm 60mm]{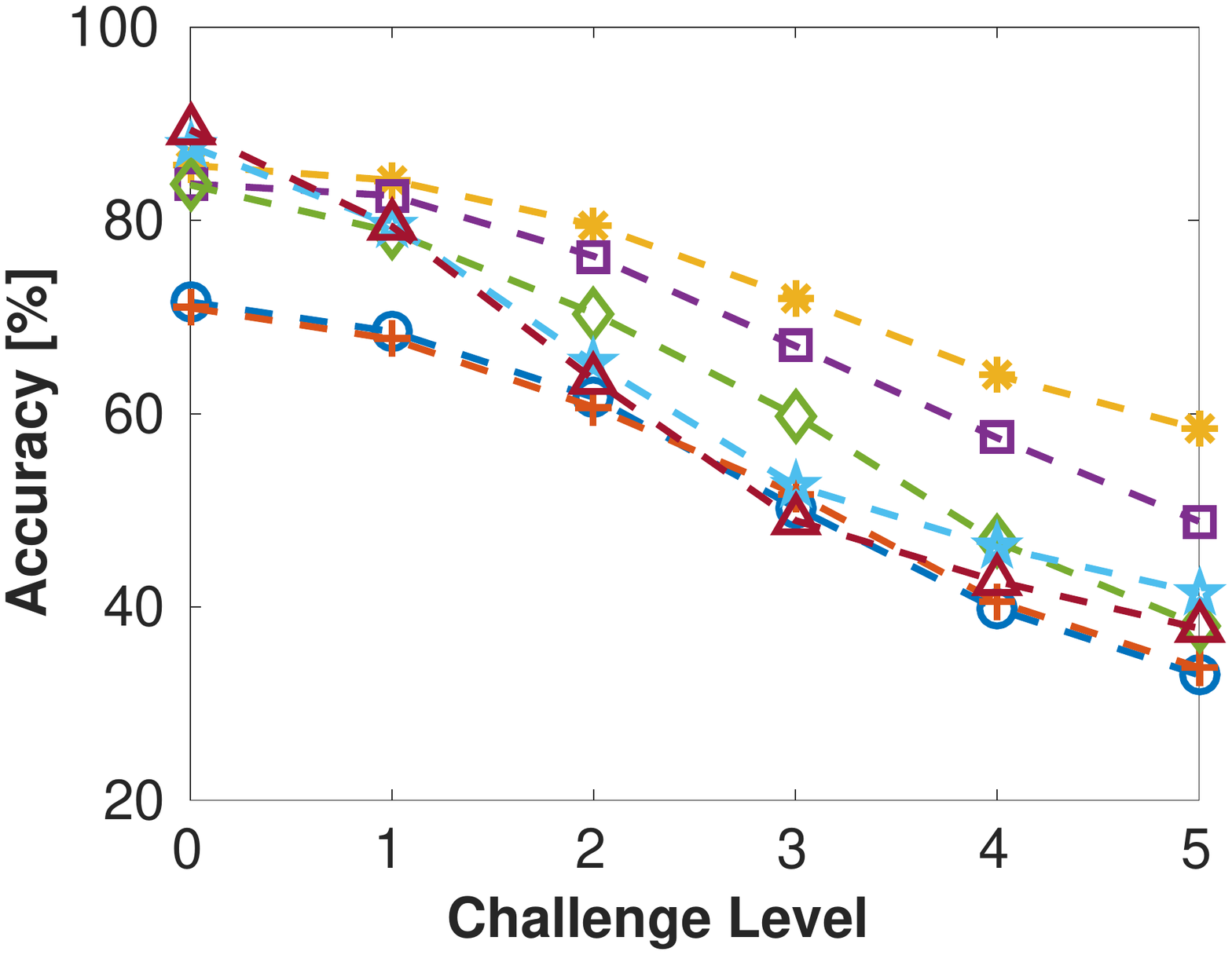}
  \vspace{0.01cm}
  \centerline{\footnotesize{(b) Lens Blur}}
\end{minipage}
\hfill
\begin{minipage}[b]{0.23\linewidth}
  \centering
\includegraphics[width=\linewidth, trim = 20mm 60mm 20mm 60mm]{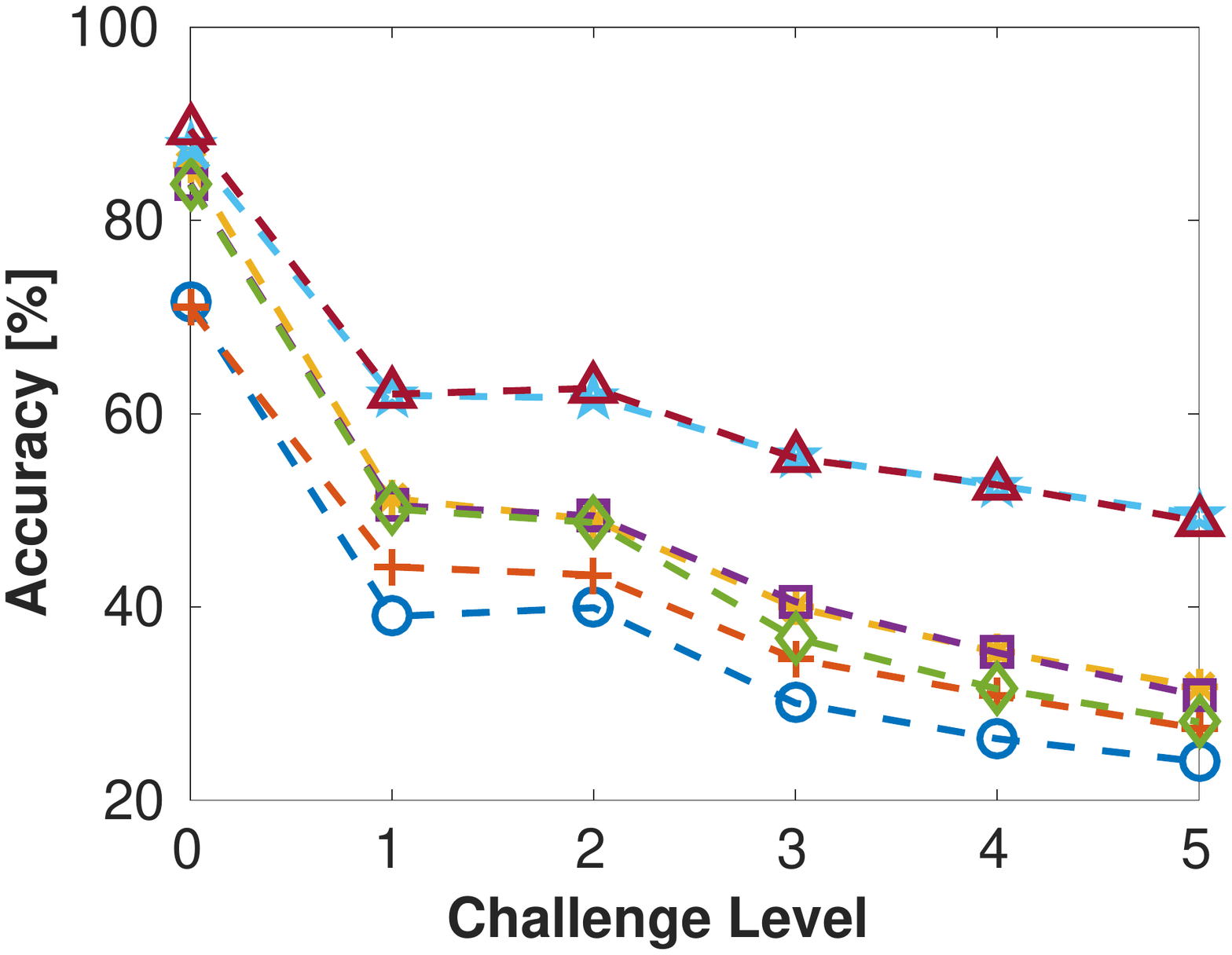}
  \vspace{0.01 cm}
  \centerline{\footnotesize{(c) Codec Error} }
\end{minipage}
\hfill
\begin{minipage}[b]{0.23\linewidth}
  \centering
\includegraphics[width=\linewidth, trim = 20mm 60mm 20mm 60mm]{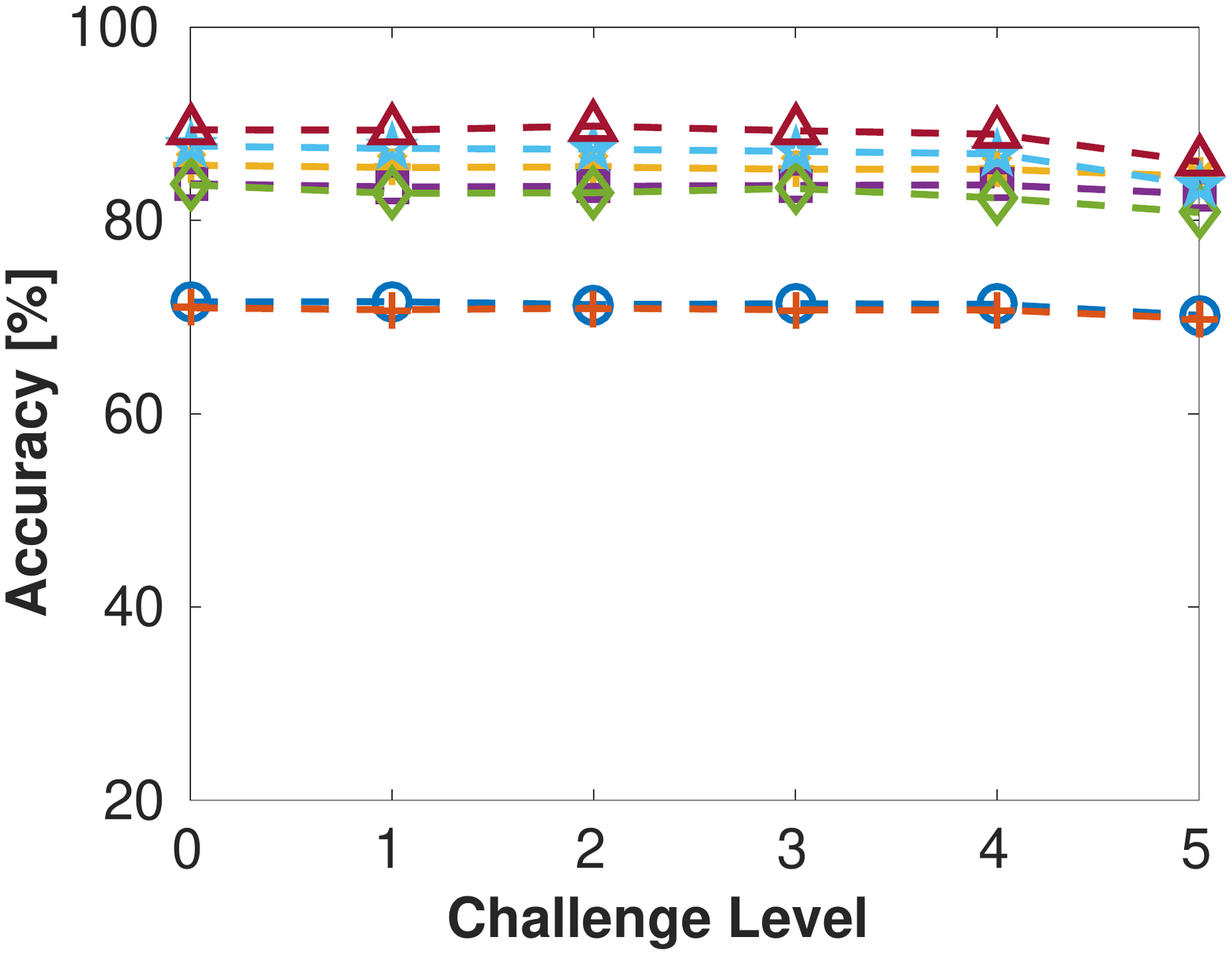}
\vspace{0.01 cm}
  \centerline{\footnotesize{(d) Darkening   } }
\end{minipage}
\hfill
\begin{minipage}[b]{0.23\linewidth}
  \centering
\includegraphics[width=\linewidth, trim = 20mm 60mm 20mm 60mm]{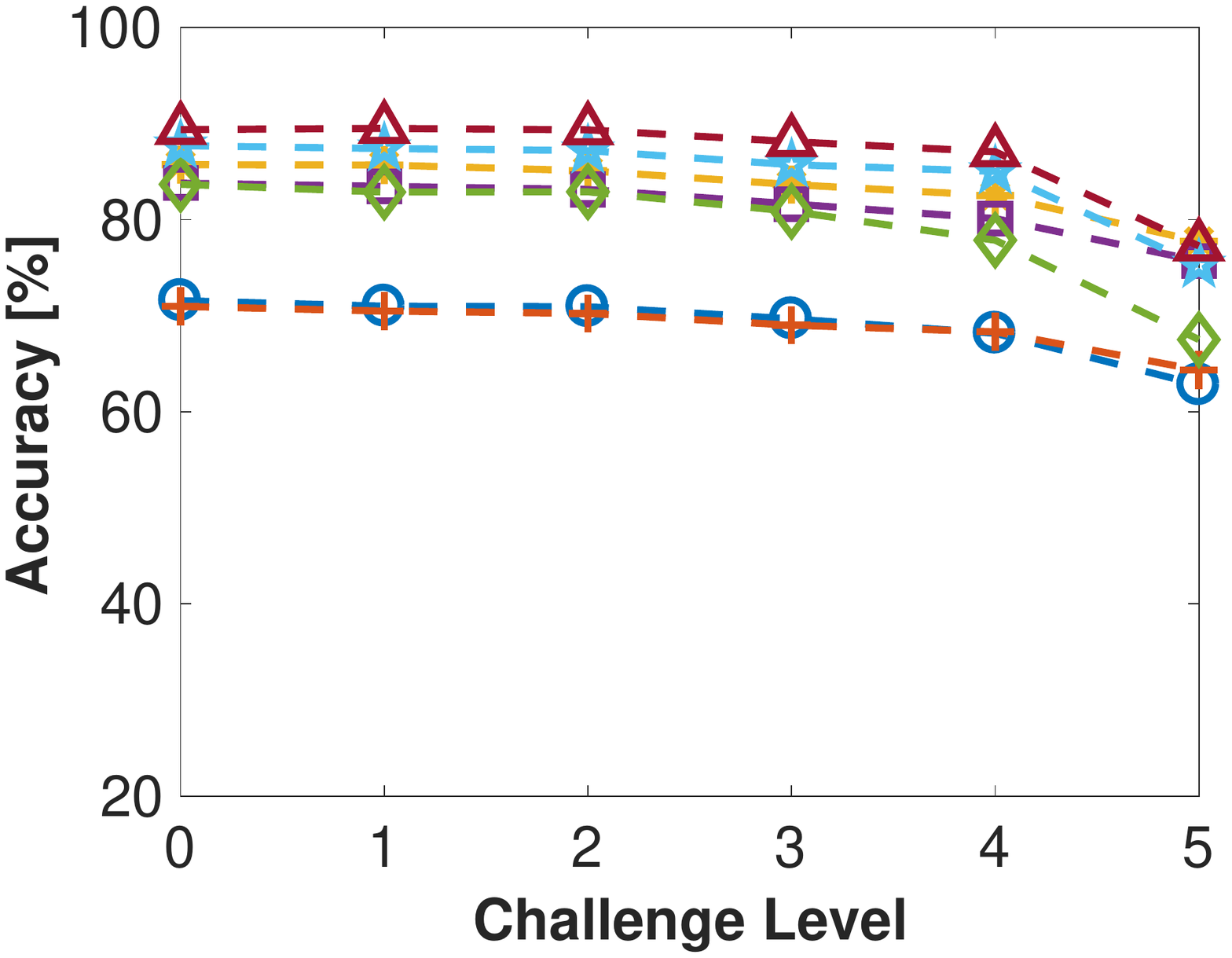}
\vspace{0.01 cm}
  \centerline{\footnotesize{(e) Dirty Lens   } }
\end{minipage}
\hfill
\begin{minipage}[b]{0.23\linewidth}
  \centering
\includegraphics[width=\linewidth, trim = 20mm 60mm 20mm 60mm]{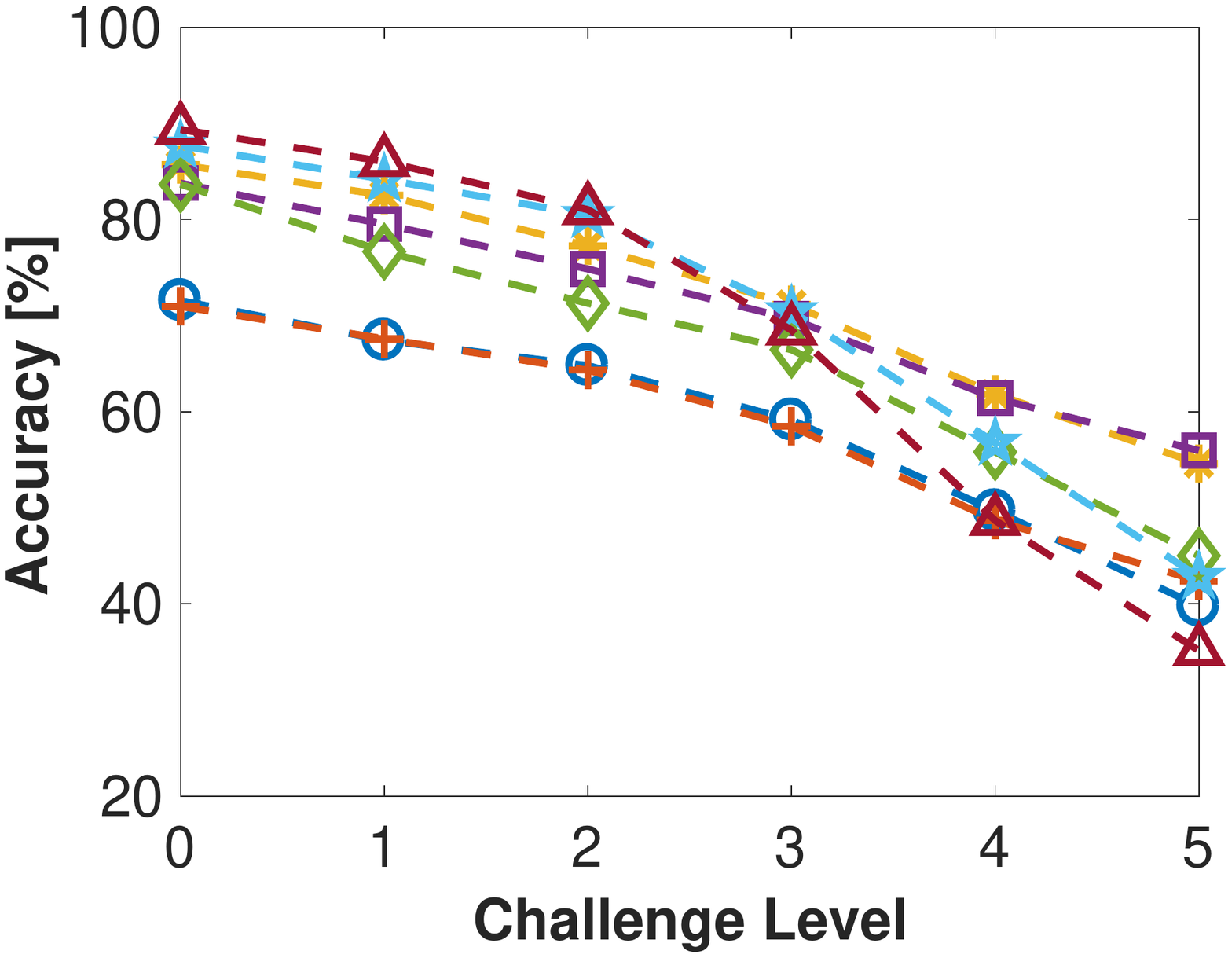}

  \vspace{0.01 cm}
  \centerline{\footnotesize{(f) Exposure  } }
\end{minipage}
\hfill
\begin{minipage}[b]{0.23\linewidth}
  \centering
\includegraphics[width=\linewidth, trim = 20mm 60mm 20mm 60mm]{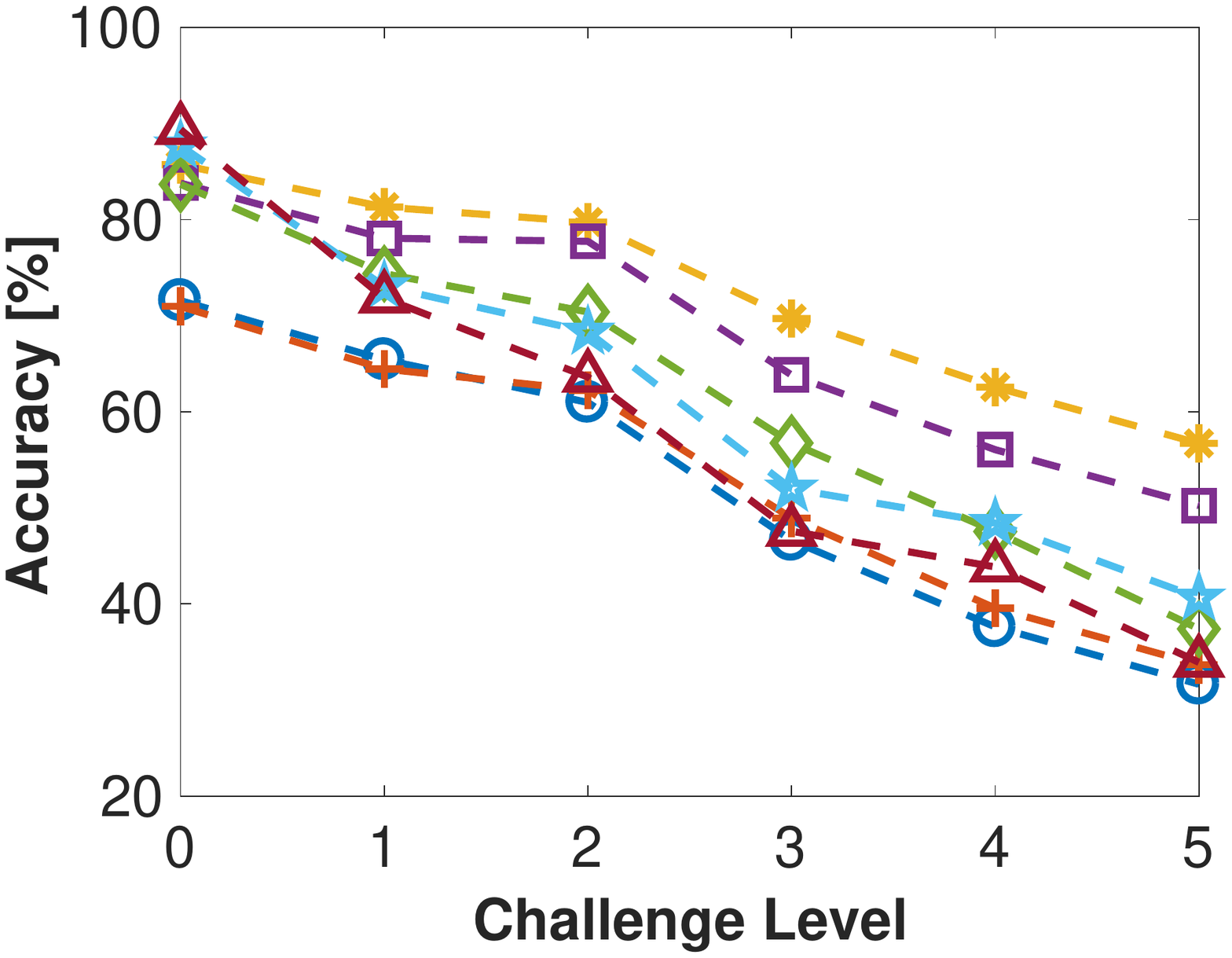}

  \vspace{0.01 cm}
  \centerline{\footnotesize{(g) Gaussian Blur   } }
\end{minipage}
\hfill
\begin{minipage}[b]{0.23\linewidth}
  \centering
\includegraphics[width=\linewidth, trim = 20mm 60mm 20mm 60mm]{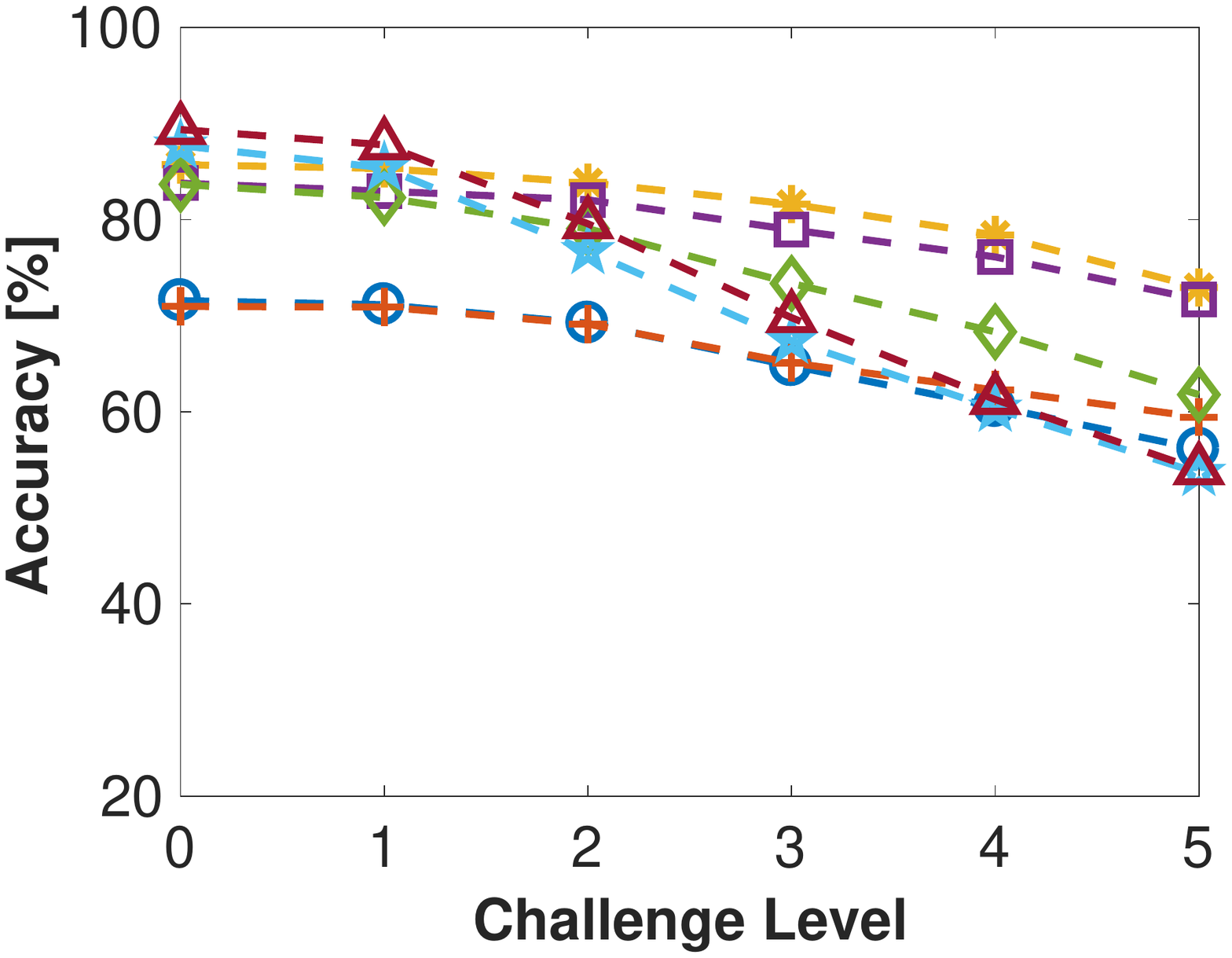}

  \vspace{0.01 cm}
  \centerline{\footnotesize{(h) Noise   } }
\end{minipage}
\hfill
\begin{minipage}[b]{0.23\linewidth}
  \centering
\includegraphics[width=\linewidth, trim = 20mm 60mm 20mm 60mm]{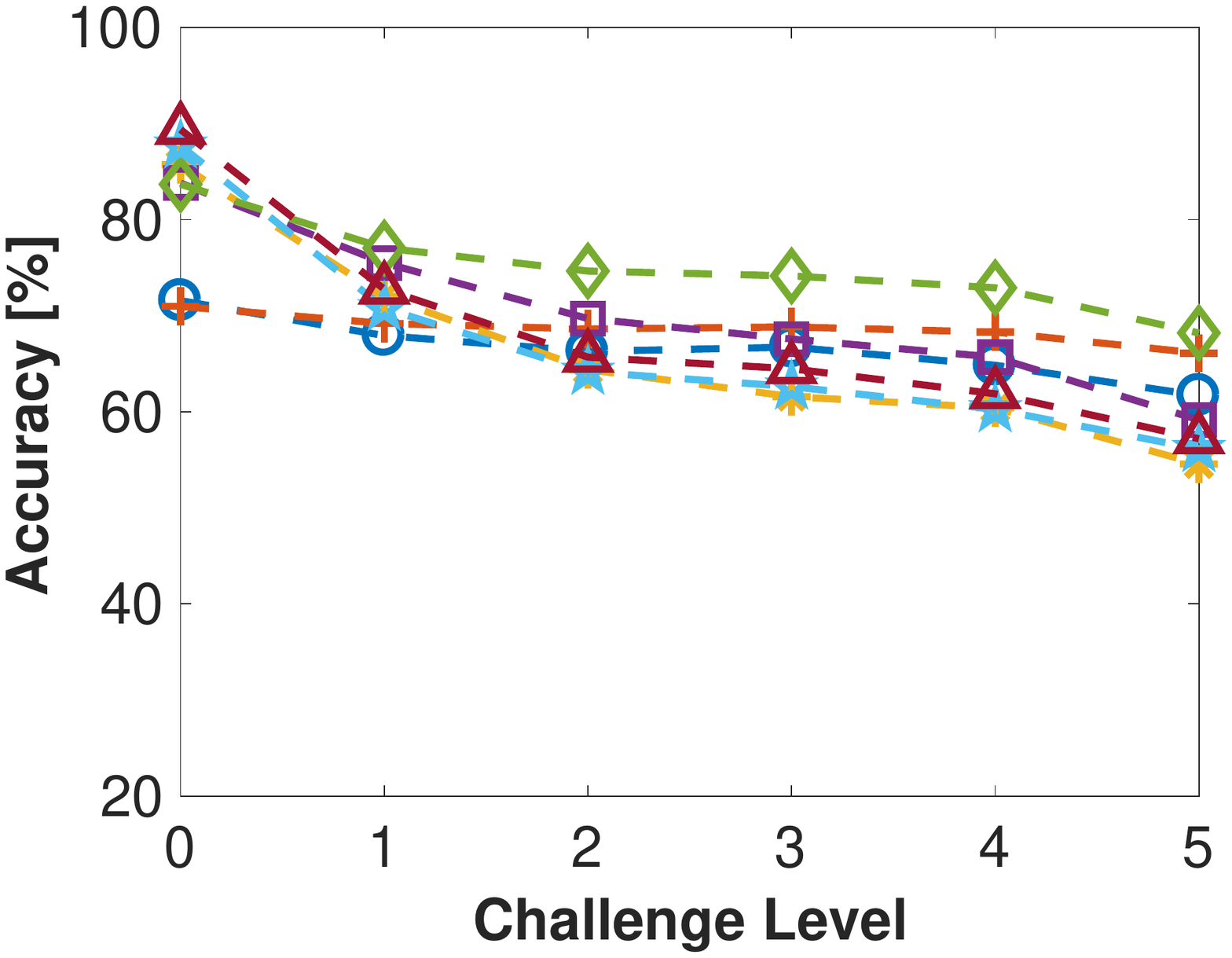}

  \vspace{0.01 cm}
  \centerline{\footnotesize{(i) Rain  } }
\end{minipage}
\hfill
\begin{minipage}[b]{0.23\linewidth}
  \centering
\includegraphics[width=\linewidth, trim = 20mm 60mm 20mm 60mm]{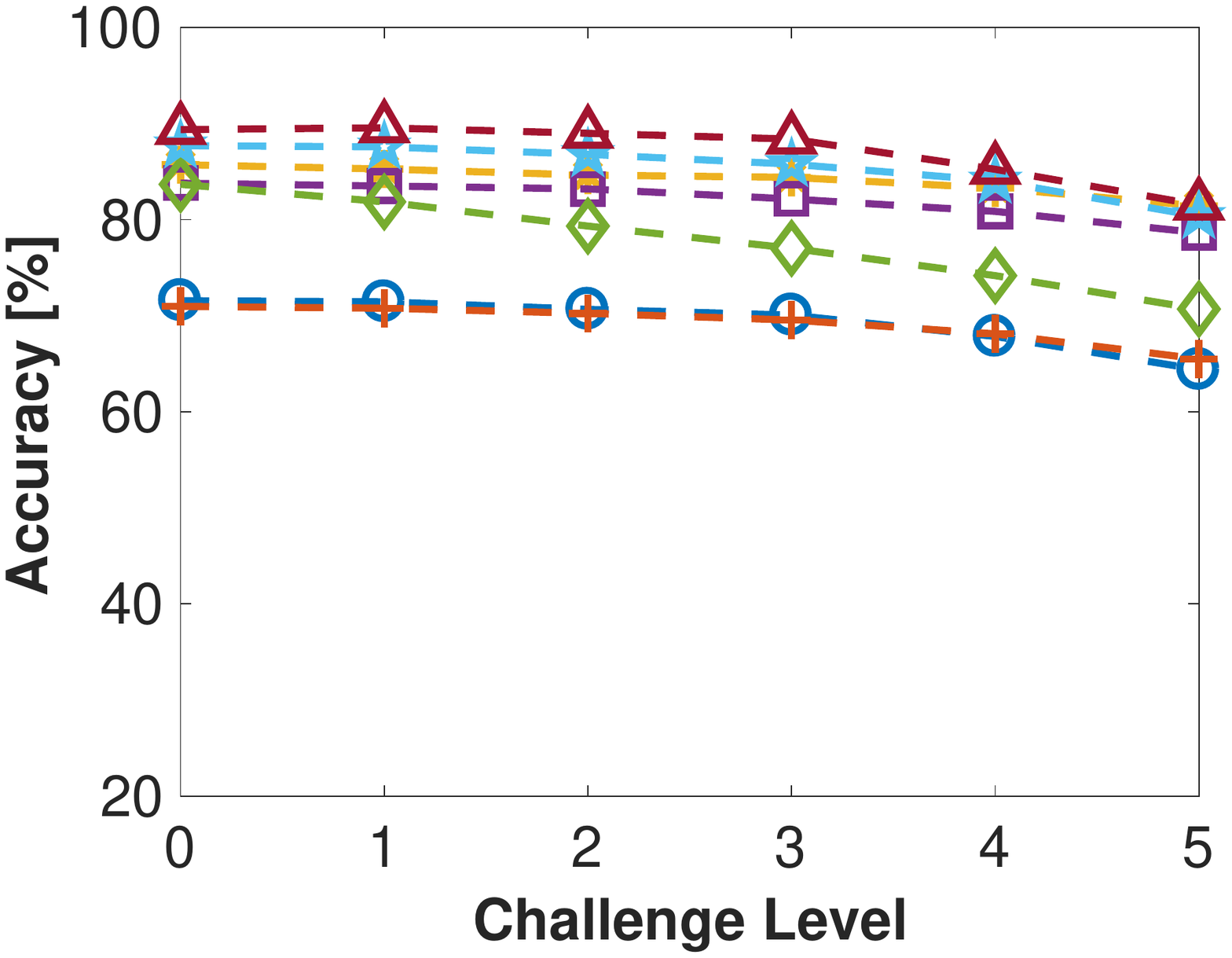}

  \vspace{0.01 cm}
  \centerline{\footnotesize{(j) Shadow   } }
\end{minipage}
\hfill
\begin{minipage}[b]{0.23\linewidth}
  \centering
\includegraphics[width=\linewidth, trim = 20mm 60mm 20mm 60mm]{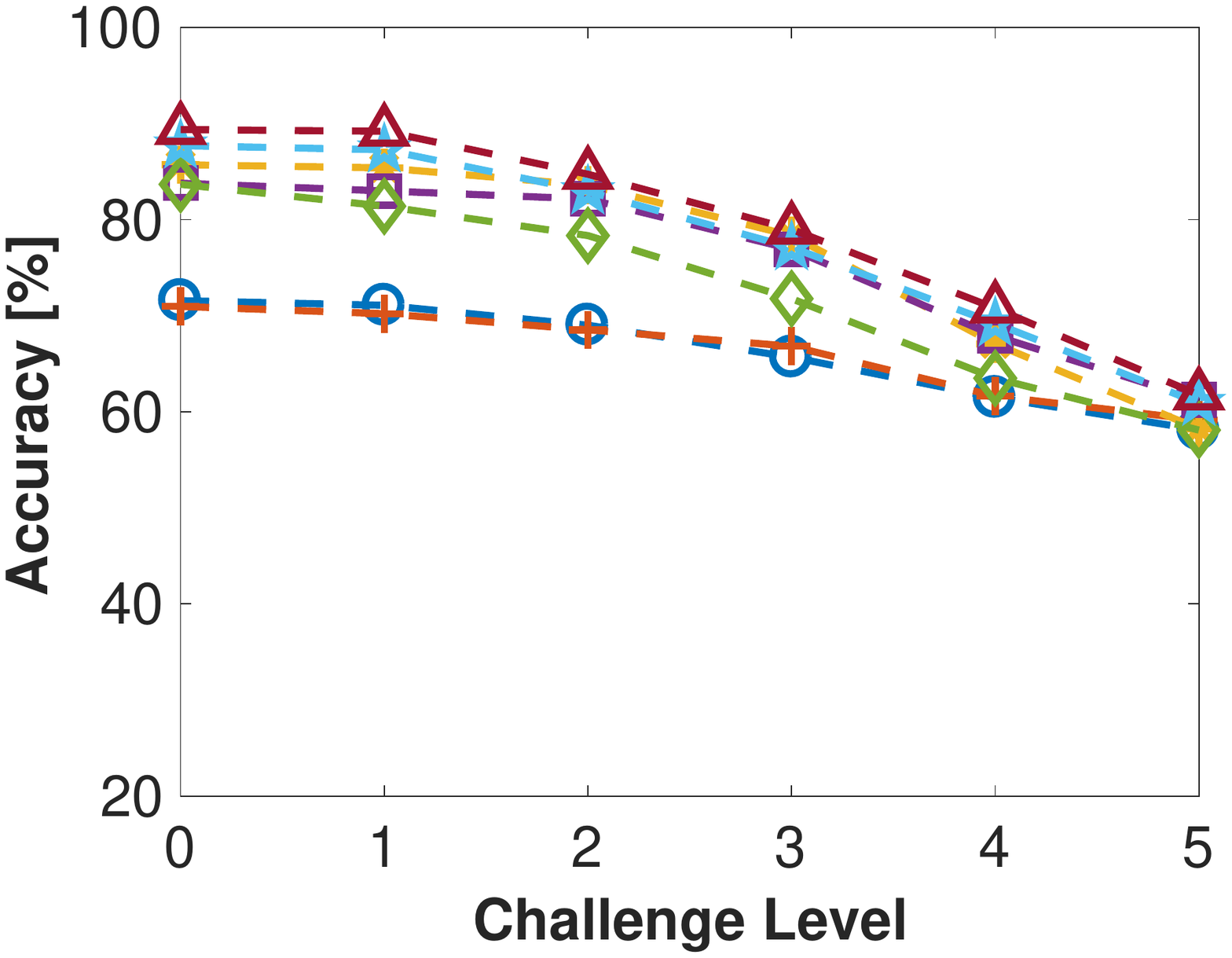}

  \vspace{0.01 cm}
  \centerline{\footnotesize{(k) Snow   } }
\end{minipage}
\hfill
\begin{minipage}[b]{0.23\linewidth}
  \centering
\includegraphics[width=\linewidth, trim = 20mm 60mm 20mm 60mm]{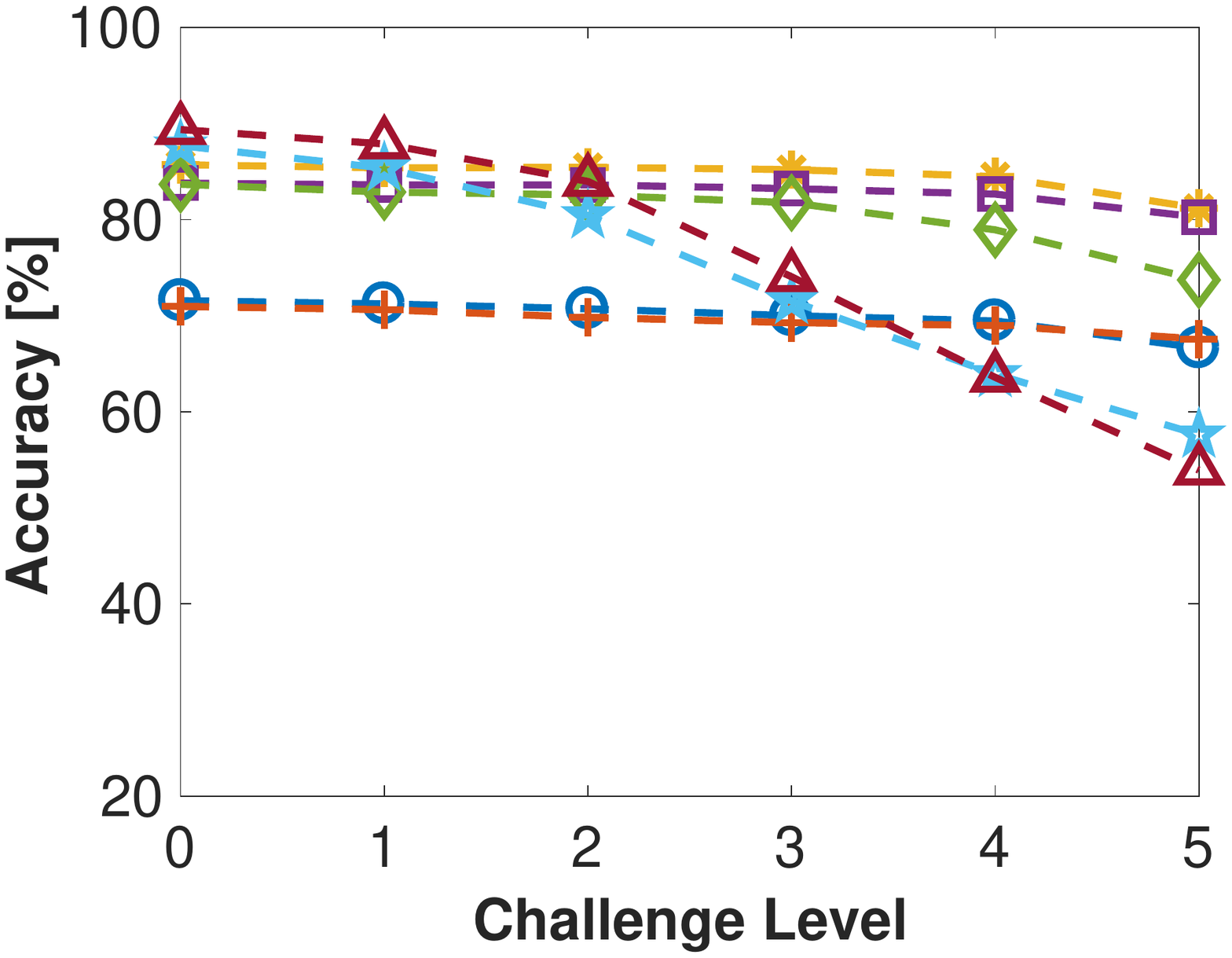}

  \vspace{0.01 cm}
  \centerline{\footnotesize{(l) Haze  } }
\end{minipage}

\vspace{0.3cm}
\centering
\includegraphics[width=\linewidth]{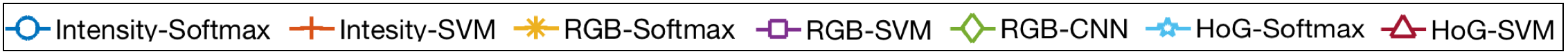}

\caption{Performance versus challenge levels.}
\vspace{-5.0mm}
\label{fig:performanceVsLevel}
\end{figure}

\end{center}

We analyze the accuracy of baseline methods with respect to challenge levels for each challenge type and report the results in Fig. \ref{fig:performanceVsLevel}. Severe \texttt{decolorization} (Fig.~\ref{fig:performanceVsLevel}(a)) leads to at least $10\%$ decrease in accuracy for color-based and HoG-based methods. However, intensity-based methods show consistent performance over different challenge levels since no color information is used by intensity-based methods. Among all the challenges, \texttt{codec error} is the most effective category that significantly degrades the classification accuracy even with challenge level $1$ as shown in Fig.~\ref{fig:performanceVsLevel}(c). We can observe that there is at least $30\%$ decrease for each method after challenge level $1$ and at least $46\%$ decrease after challenge level $5$. \texttt{Lens blur} (Fig.~\ref{fig:performanceVsLevel}(b)), \texttt{exposure} (Fig.~\ref{fig:performanceVsLevel}(f)), and \texttt{Gaussian blur} (Fig.~\ref{fig:performanceVsLevel}(g)) result in significant performance decrease under severe challenging conditions, at least $27\%$ for each baseline method. However, classification accuracy decreases more linearly in these categories compared to \texttt{codec error} because of its steep decrease in level $1$. In \texttt{darkening} category (Fig.~\ref{fig:performanceVsLevel}(d)), classification accuracy is consistent for all the methods. The normalization operation in the preprocessing step makes all methods less sensitive to \texttt{darkening} challenge. When challenge level becomes more severe, performance of baseline methods degrades a few percent at most.

In \texttt{dirty lens} category (Fig.~\ref{fig:performanceVsLevel}(e)), new dirty lens images were overlaid on entire images to increase the challenge level. The new dirt patterns do not necessarily occlude traffic signs. Therefore, performance of baseline methods do not always change when challenge level increases. In \texttt{noise} category (Fig.~\ref{fig:performanceVsLevel}(h)), HoG and CNN correspond to a more linear performance decrease compared to intensity and color-based methods. In \texttt{rain} category (Fig.~\ref{fig:performanceVsLevel}(i)), particle models are all around the scene, which result in significant occlusion even in level $1$ challenge. Therefore, degradation while going from challenge-free to level $1$ challenge is steeper than any further relative changes for color-based method, HoG-based method, and CNN. In \texttt{shadow} category (Fig.~\ref{fig:performanceVsLevel}(j)), vertical shadow lines are all over the images. We observe slight degradation as challenge level increases because areas under shadow become less visible. In case of \texttt{snow} challenge (Fig.~\ref{fig:performanceVsLevel}(k)), all methods converge to a similar classification accuracy under severe \texttt{snow} challenge. In \texttt{haze} category (Fig.~\ref{fig:performanceVsLevel}(l)), performance of intensity-based, color-based, and CNN methods is relatively consistent whereas decrease in HoG-based models follows a more linear behavior. Color image-based classifiers and CNN are less sensitive to \texttt{haze} challenge compared to other methods. \texttt{Haze} challenge was generated as a combination of radial gradient operator with partial opacity, a smoothing operator, an exposure operator, a brightness operator, and a contrast operator. Moreover, the location of the operator was adjusted manually per frame to simulate a sense of depth. Because of the complexity of \texttt{haze} model, it is less intuitive to explain the behavior of baseline methods. However, the higher tolerance of CNN model with respect to \texttt{haze} challenge can be explained with its capability to directly learn spatial patterns from visual representations.  


\subsection{Experiment 2: Recognition in Real Environments under Challenging Conditions with Data Augmentation}
We investigate the role of data augmentation methods in traffic sign recognition under challenging conditions. Augmented data include flipped real images, real challenge images, and unreal challenge images. To augment flipped images, real traffic sign images were randomly flipped horizontally, vertically or horizontally and vertically. To augment real challenge images, we selected 20 traffic sign images with maximum area (highest resolution samples) for each traffic sign in the training set. Then, we obtained corresponding images for each challenge type and level. It should be noted that augmented data is challenge version of the challenge-free data, which is already utilized in the training. We utilized the same source images because challenging conditions in the original video sequences were synthesized globally over entire videos and it is not possible to apply the same challenge generation framework directly over new traffic sign images. Overall, in both augmentation experiments, training set include $3,080$ images ($20$ images $\times 11$ challenge types $\times 14$ traffic signs) and $7,292$ (complete challenge-free training set) real images. Test set is same as experiment $1$ for all three data augmentation methods. 

Data augmentation based on flipping degrades the recognition performance by $6.78\%$. Decrease in recognition performance can be mainly because of asymmetric characteristics of traffic sign images. We further explain details about this performance decrease in Appendix~\ref{sec:AppendixB}. Data augmentation with real challenge images slightly decreases the average performance by $0.45\%$. Even though novel challenging conditions are added in the augmentation stage, original images are already included in the training. Therefore, such augmentation method does not lead to any performance enhancement in tested scenarios. Similar to real challenge images, we obtained unreal challenge images by selecting the traffic signs with maximum area for each traffic sign, challenge type, and challenge level. Instead of using solely 20 distinct images for each sign, we utilize 220 distinct unreal images (one distinct image for each sign and challenge category) in the data augmentation, which enriches training set with new angle, contrast, and lighting configurations. Results of unreal image-based data augmentations are summarized in Table \ref{tab_results}. Each entry in the table other than the last row and the last column was obtained by calculating the performance change for a baseline method over all the challenge levels for a specific challenge type. Entries in the last row  were calculated by averaging the performance change of each baseline method over all challenge types. Finally, entries in the last column were calculated by averaging the performance change  over all baseline methods for each challenge type.



%

\begin{table}[htbp!]
\footnotesize
\centering
\caption{Classification accuracy change (\%) when additional unreal images used in the training. }
\label{tab_results}
\begin{tabular}{ccccccccc}
\hline
\multicolumn{1}{l}{\multirow{3}{*}{\textbf{Challenge Types}}} & \multicolumn{7}{c}{\textbf{Baseline Methods}}                                                                                                                                \\ \cline{2-9} 
\multicolumn{1}{l}{}                                          & \multicolumn{2}{c}{\textbf{Intensity}}       & \multicolumn{2}{c}{\textbf{Color}}           & \multicolumn{2}{c}{\textbf{HoG}}             & \multirow{2}{*}{\textbf{CNN}} & \multirow{2}{*}{\textbf{Average}}\\ \cline{2-7}                              & \textbf{Softmax}      & \textbf{SVM}          & \textbf{Softmax}      & \textbf{SVM}          & \textbf{Softmax}      & \textbf{SVM}          &                               \\ \hline
\textbf{Decolorization} & \bf +2.86 & \bf +3.32 & \bf +1.46 & -0.53 & \bf +1.43 & -0.01 & \bf +3.23 & \bf +1.68\\ 
\textbf{Lens Blur} & \bf +3.98 & \bf +2.71 & \bf +4.45 & \bf +6.60 & \bf +3.34 & \bf +1.81 & -1.78 & \bf +3.02\\ 
\textbf{Codec Error} & \bf +0.47 & -1.21 & \bf +1.51 & -0.82 & -1.55 & -1.61 & \bf +2.40 & -0.12\\ 
\textbf{Darkening} & \bf +2.83 & \bf +2.98 & \bf +2.87 & \bf +1.44 & \bf +1.68 & \bf +0.44 & \bf +2.58 & \bf +2.12\\ 
\textbf{Dirty lens} & \bf +3.14 & \bf +2.86 & \bf +2.68 & \bf +1.63 & \bf +2.00 & \bf +0.62 &  \bf +3.11 & \bf +2.29\\ 
\textbf{Exposure} & \bf +2.54 & \bf +1.77 & \bf +1.34 & \bf +1.97 & -0.66 & -2.23 & \bf +0.54 & \bf +0.75\\ 
\textbf{Gaussian Blur} & \bf +5.89 & \bf +3.98 & \bf +4.24 & \bf +7.06 & \bf +2.03 & \bf +1.77 & \bf +2.78 & \bf +3.97\\ 
\textbf{Noise} & \bf +1.62 & \bf +1.58 & \bf +1.89 & \bf +0.58 & \bf +1.41 & -0.90 & \bf +2.25 & \bf +1.21 \\ 
\textbf{Rain}  & \bf +2.30 & \bf +1.28 & \bf +4.73 & \bf +2.75 & \bf +5.48 & \bf +2.34 & \bf +0.69 & \bf +2.80 \\ 
\textbf{Shadow}     & \bf +2.95 & \bf +3.38 & \bf +3.27 & \bf +1.62 & \bf +1.73 & \bf +0.64 & \bf +3.01 & \bf +2.37 \\ 
\textbf{Snow} & \bf +3.19 & \bf +2.81 & \bf +2.09 & \bf +0.48 & \bf +2.63 & \bf +0.92 & \bf +4.34 & \bf +2.35 \\
\textbf{Haze}   & \bf +3.28 & \bf +3.22 & \bf +3.22 & \bf +1.41 & \bf +2.26 & -1.35 & \bf +3.51 & \bf +2.22 \\

  \bf All (average) & \bf +2.92  & \bf +2.39 & \bf +2.81  & \bf +2.02 & \bf +1.81  & \bf +0.20  & \bf +2.22 & -\\ \hline

\end{tabular}
\end{table}

We test $7$ baseline methods over $12$ challenge types and report the performance change of each baseline method for each challenge type. Out of $84$ result categories ($7$ baseline methods $\times 12$ challenge types), classification performance increases in $72$ of them. On average, classification performance  increases for all challenge types other than a slight decrease in \texttt{codec error}. Moreover, average classification performance increases for each baseline method, which is a slight increase for HoG-SVM  ($0.2\%$) and more for other methods (at least $1.81\%$). Additional unreal images in the training set were obtained from all the challenge types except \texttt{haze} category. However, classification accuracy increases for all the baseline methods at least $1.41\%$ other than HoG-SVM in \texttt{haze} category. The performance enhancement in \texttt{haze} can be understood by analyzing the computational model of \texttt{haze} and its perceptual similarity to other challenges. \texttt{Haze} model includes a smoothing operator, an exposure filter, a brightness operator, and a contrast operator. Exposure filter is used in the \texttt{exposure} (overexposure) model and smoothing operator is utilized in \texttt{blur} models. Moreover, perceptually, we can observe similarities between \texttt{haze} and \texttt{blur} challenges in terms of smoothness and similarities between \texttt{haze} and \texttt{exposure} in terms of washed out details. Therefore, perceptually and computationally similar challenges in the training stage can affect the performance of each other in the testing stage.

\section{Conclusion}
We introduced the \texttt{CURE-TSR} dataset, which is the most comprehensive traffic sign recognition dataset in the literature that includes controlled challenging conditions. We provided a benchmark of commonly used methods in the \texttt{CURE-TSR} dataset and reported that challenging conditions lead to severe performance degradation for all baseline methods. We have shown that \texttt{lens blur}, \texttt{exposure}, \texttt{Gaussian blur}, and \texttt{codec error} degrade recognition performance more significantly compared to other challenge types because these challenge categories directly result in losing or misplacing structural information. We also investigated the effect of data augmentation and showed that flipping or simply adding challenging conditions to training data do not necessarily enhance recognition performance. However, experimental results showed that utilization of diverse simulator data with challenging conditions can enhance the average recognition performance in real-world scenarios.

\newpage
  



\newpage

\appendix
\section{Appendix: Visualization of Challenge Levels and Types}
\label{sec:AppendixA}
\begin{figure}[h]
    \centering
    \includegraphics{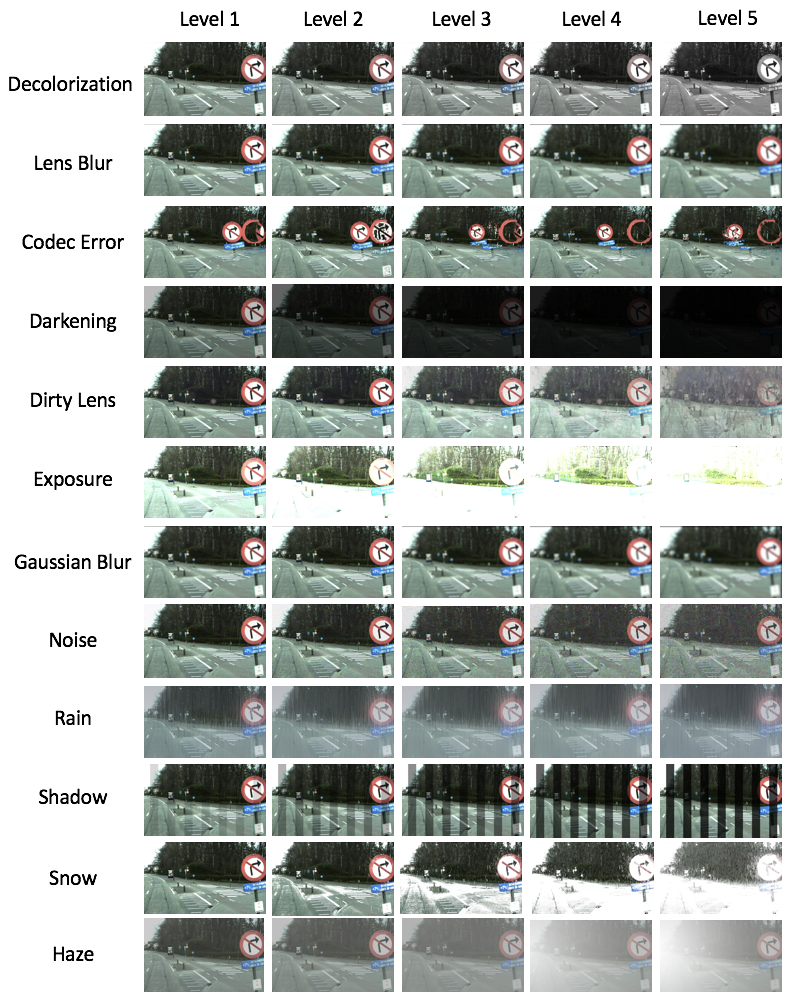}
    \caption{Challenging scene examples from each challenge type and level in CURE-TSD \cite{curetsd_dataset} and CURE-TSR \cite{curetsr_dataset} datasets.}
    \label{fig:cure-tsd-all}
\end{figure}
To visualize scenes with realistic challenging condition types and levels, we cropped surrounding environments with traffic signs as shown in Fig.~\ref{fig:cure-tsd-all}. Each row corresponds to a challenging condition and each column corresponds to a certain level of the challenging condition. Compared to other existing datasets, the \texttt{CURE-TSR} dataset contains more diverse challenging conditions and levels, which enables a comprehensive platform to test the robustness of recognition algorithms under challenging conditions. 

\section{Appendix: Data Augmentation}
\label{sec:AppendixB}
We retrained the benchmark algorithms, listed in Sec. \ref{exp_setup}, by augmenting the initial training images with their flipped versions. Vertically, horizontally or vertically and horizontally flipped challenge-free real images were used for data augmentation. Translation was not utilized in the data augmentation because recognition dataset is based on cropped images that do not include background information. Flipping-based data augmentation degrades the average recognition accuracy by more than $6.5\%$ mainly because of the asymmetric characteristics of traffic sign images. For instance, consider the \textit{no stopping} and \textit{no parking} signs from Fig. \ref{fig: sign_type}. The former sign is symmetric along its horizontal axis while the latter sign is asymmetric. Augmenting the training data with horizontally flipped versions of the asymmetric \textit{no parking} can lead to learning a visual representation similar to  \textit{no stopping} sign, which is different from the intended class and can degrade the overall degradation accuracy.     

\begin{figure}[htbp!]
\centering
\begin{minipage}[b]{0.30\linewidth}
\includegraphics[width=\linewidth]{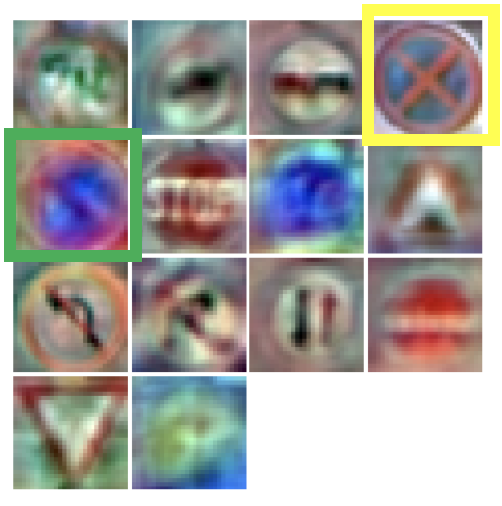}
  \vspace{0.05cm}
  \centerline{\footnotesize{(a) Softmax model }}
\end{minipage}
\hspace{1cm}
\begin{minipage}[b]{0.40\linewidth}
\includegraphics[width=\linewidth]{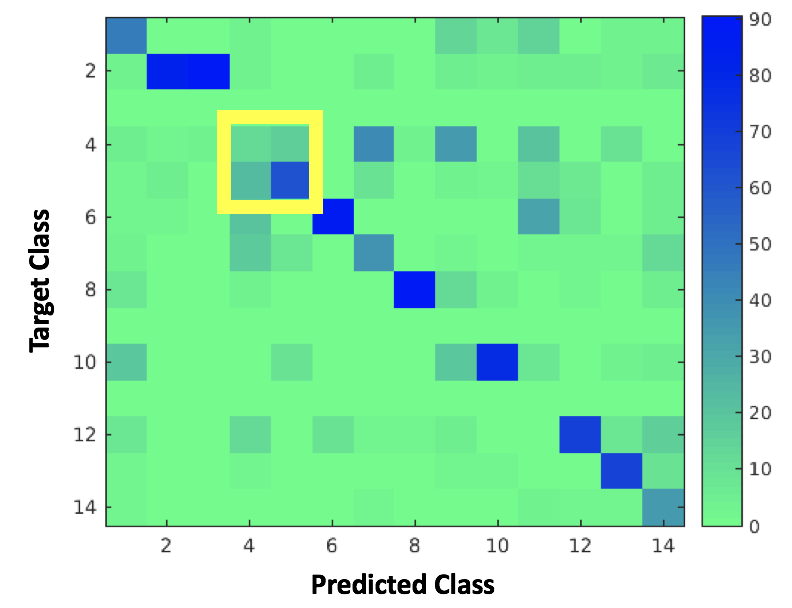}
  \vspace{0.05cm}
  \centerline{\footnotesize{\vspace{5cm}(b) Confusion Matrix}}
\end{minipage}
\
\caption{RGB softmax model visualization and averaged confusion matrix without data augmentation.}
\label{fig:WithoutDA}
\end{figure}
\begin{figure}[htbp!]
\centering
\begin{minipage}[b]{0.30\linewidth}
\includegraphics[width=\linewidth]{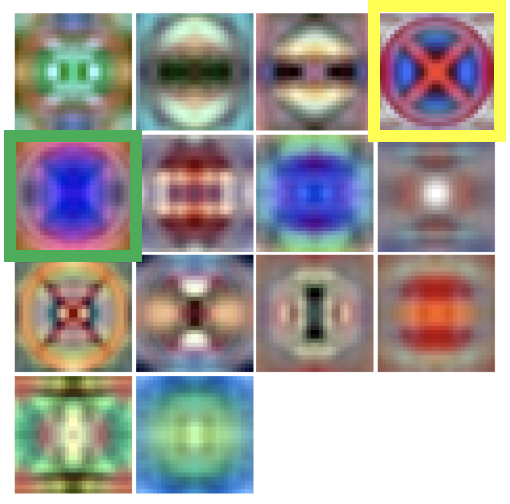}
  \vspace{0.05cm}
  \centerline{\footnotesize{(a) Softmax model }}
\end{minipage}
\hspace{1cm}
\begin{minipage}[b]{0.40\linewidth}
\includegraphics[width=\linewidth]{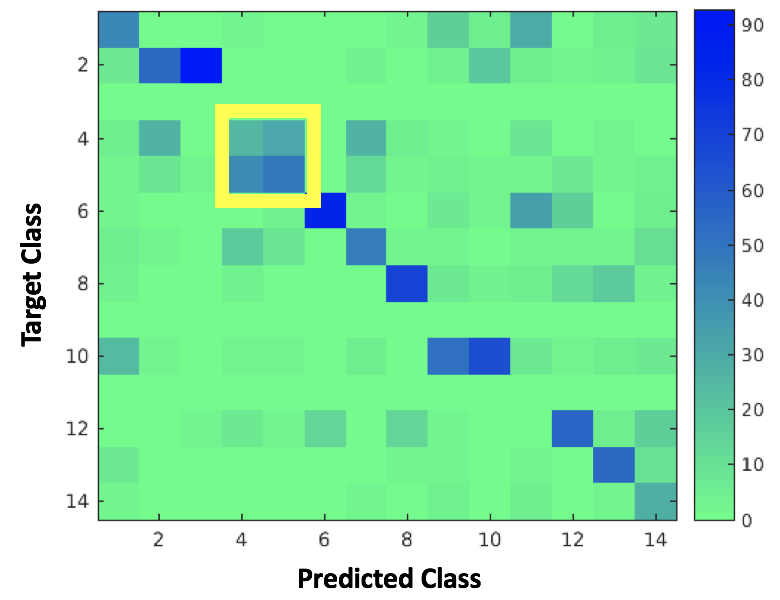}
  \vspace{0.05cm}
  \centerline{\footnotesize{(b) Confusion matrix}}
\end{minipage}
\
\caption{RGB softmax model visualization and averaged confusion matrix with data augmentation.}
\label{fig:WithDA}
\end{figure}

We visualize two softmax RGB trained models, one of which was trained without data augmentation while the other was trained with data augmentation (flipped real images) in Figs. \ref{fig:WithoutDA} and \ref{fig:WithDA} respectively. Consider the case of the learned \textit{no parking} model ($2^{nd}$ row, $1^{st}$ column, green highlight). Perceptually, the data-augmented model has two diagonal lines crossing each other as opposed to the single diagonal in the model learned without data augmentation. However, lines that perpendicularly cross is a characteristic of the learned \textit{no stopping} model ($1^{st}$ row, $4^{th}$ column, yellow highlight), which would result in misclassification. Increase in the misclassification rate between these two signs because of flipping-based data augmentation can be understood from the confusion matrices in Fig.\ref{fig:WithoutDA}(b) and Fig.\ref{fig:WithDA}(b) (highlighted in yellow: class types 4 and 5) in which darker colors correspond to more misclassification.

\end{document}